\documentclass[runningheads]{llncs}

\usepackage{eccv}

\usepackage{eccvabbrv}

\usepackage{graphicx}
\usepackage{booktabs}

\usepackage[accsupp]{axessibility}  

\usepackage{hyperref}

\usepackage{orcidlink}

\usepackage{bbm}
\usepackage{makecell}
\usepackage{marvosym}
\usepackage{multirow}
\usepackage{arydshln}
\usepackage[ruled]{algorithm2e}

\begin{document}

\title{Strike a Balance in \\
Continual Panoptic Segmentation} 


\author{Jinpeng Chen\inst{1}\orcidlink{0000-0002-0469-4463} \and
Runmin Cong\inst{2,3}$^{(\textrm{\Letter})}$\orcidlink{0000-0003-0972-4008} \and
Yuxuan Luo\inst{1}\orcidlink{0000-0003-1003-2252} \and \\
Horace Ho Shing Ip\inst{1,4}\orcidlink{0000-0002-1509-9002} \and
Sam Kwong\inst{5}$^{(\textrm{\Letter})}$\orcidlink{0000-0001-7484-7261}}


\authorrunning{J.~Chen et al.}

\institute{Department of Computer Science, City University of Hong Kong, Hong Kong \\ \email{\{jinpeng.chen,yuxuanluo4-c\}@my.cityu.edu.hk} \\ \email{horace.ip@cityu.edu.hk} \and
School of Control Science and Engineering, Shandong University, \\Jinan, Shandong, China \\ \email{rmcong@sdu.edu.cn} \and
Key Laboratory of Machine Intelligence and System Control, \\Ministry of Education, Jinan, Shandong, China \and
Centre for Innovative Applications of Internet and Multimedia Technologies, \\City University of Hong Kong, Hong Kong \and
Lingnan University, Hong Kong \\ \email{samkwong@ln.edu.hk}}


\maketitle


\begin{abstract}
  This study explores the emerging area of continual panoptic segmentation, highlighting three key balances. First, we introduce past-class backtrace distillation to balance the stability of existing knowledge with the adaptability to new information. This technique retraces the features associated with past classes based on the final label assignment results, performing knowledge distillation targeting these specific features from the previous model while allowing other features to flexibly adapt to new information. Additionally, we introduce a class-proportional memory strategy, which aligns the class distribution in the replay sample set with that of the historical training data. This strategy maintains a balanced class representation during replay, enhancing the utility of the limited-capacity replay sample set in recalling prior classes. Moreover, recognizing that replay samples are annotated only for the classes of their original step, we devise balanced anti-misguidance losses, which combat the impact of incomplete annotations without incurring classification bias. Building upon these innovations, we present a new method named Balanced Continual Panoptic Segmentation (BalConpas). Our evaluation on the challenging ADE20K dataset demonstrates its superior performance compared to existing state-of-the-art methods. The official code is available at \url{https://github.com/jinpeng0528/BalConpas}
\end{abstract}

\section{Introduction}
\label{sec:intro}

Panoptic segmentation \cite{kirillov2019panoptic}, which integrates the concepts of semantic and instance segmentation, is a foundational task in computer vision. This task aims to classify each pixel of an image into unique semantic categories while also distinguishing between different instances. The standard practice involves training models using static datasets. However, when these datasets undergo updates, it becomes necessary to retrain the network entirely. This is because fine-tuning with only new data can lead to catastrophic forgetting of previously learned information. This limitation poses a challenge in adapting to environmental shifts or new demands. Thus, a pivotal research question is how to enable panoptic segmentation models to assimilate new information without losing previously acquired knowledge, a task known as continual panoptic segmentation (CPS).

In CPS, CoMFormer \cite{cermelli2023comformer} is a pioneering work. It addresses catastrophic forgetting by employing output-level knowledge distillation. In addition, it tackles background shift, another significant issue in continual segmentation, by using pseudo-labels generated from the previous model to supplement annotations of past classes. Background shift \cite{cermelli2020modeling} refers to the mislabeling of previously learned foreground classes as background due to the absence of their annotations, potentially overturning established knowledge. Beyond CoMFormer, techniques from continual semantic segmentation (CSS) methods, such as feature-level knowledge distillation \cite{douillard2021plop, zhang2022representation} and sample replay \cite{maracani2021recall, cha2021ssul, baek2022decomposed}, might offer additional insights for CPS. Nonetheless, we argue that striking a balance in three critical aspects is crucial, and current methods have not yet provided an optimal solution.

\begin{figure}[!t]
\centering
\includegraphics[width=0.75\textwidth]{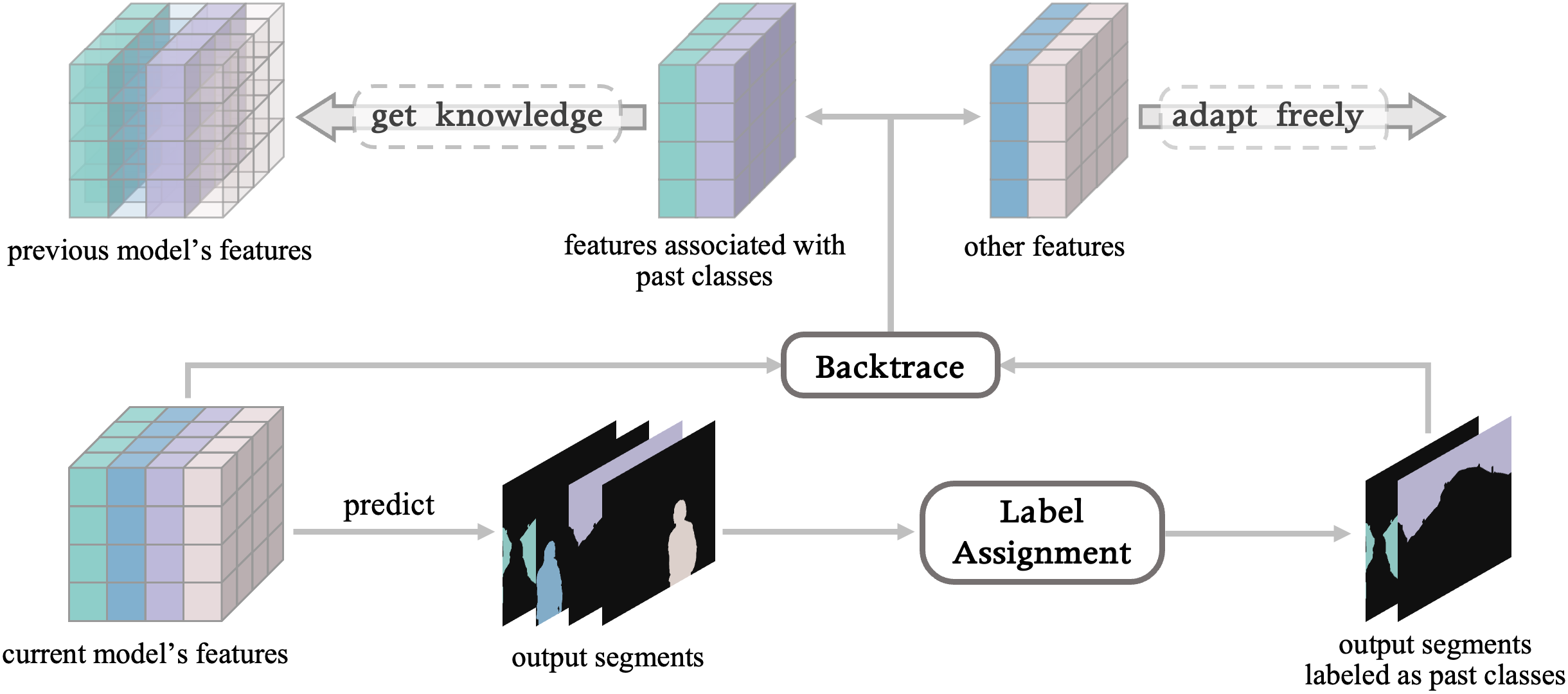}
\caption{Illustration of our past-class backtrace distillation. It retraces features associated with output segments labeled as past classes for targeted knowledge distillation, while simultaneously allowing other features to adapt freely to new knowledge.}
\label{fig:pcbd}
\end{figure}

The first balance we consider is between maintaining stability in prior knowledge and fostering adaptability for new information in knowledge distillation. Existing methods \cite{michieli2019incremental, baek2022decomposed, zhang2022representation} implement distillation from the previous model to the current one, targeting entire features or outputs. While effective in remembering past classes, this strategy can limit the learning of new ones. We propose that balancing these two needs requires selectively distilling features relevant to prior knowledge, as the previous model contains valuable information only on these features. Other features should be allowed to freely adapt to new knowledge. In segmentation tasks, a challenge arises when an input image contains both new and past classes, making it difficult to identify the features related to the latter. To address this, we introduce past-class backtrace distillation, depicted in Fig. \ref{fig:pcbd}. This process starts after the label assignment \cite{carion2020end, cheng2021per} of network outputs is completed. Using the assignment results, we can pinpoint the output segments related to past classes and then trace back to their corresponding earlier features for distillation. This ensures consistent recognition accuracy for past classes. Importantly, this selective distillation approach allows other features to remain adaptable, ensuring the unhindered learning of new classes.

The second key balance involves the class distribution within the replay sample set. Current methods \cite{cha2021ssul, baek2022decomposed} select an equal number of images for each class, which, although seemingly balanced, neglects the varying occurrence frequencies of classes in the training data. We propose that a true representation of class balance should mirror the cumulative distribution of classes across all previous training sets. Classes that are more prevalent in the training data are often more important in corresponding application contexts and exhibit a broader range of variations, making accurate memory of them both crucial and challenging. Hence, these classes should occupy a larger portion of the replay sample set. To achieve this, we propose a class-proportional memory strategy. This process begins by forming a replay sample set at the first step that reflects the class distribution of the first training set. Then, the set is updated after each step to represent the evolving cumulative class distribution.

Despite achieving a balanced replay sample set, an obstacle emerges during the replay process. As these samples are annotated solely with classes from the step at which they were collected, replaying them inevitably leads to the mislabeling of both current and other past classes as background. This mislabeling can impede the learning of current classes and exacerbate the forgetting of other past classes. To address this, we devise a pair of loss functions. The first, applied to replay samples, focuses exclusively on foreground classes, thereby avoiding the misguidance of incorrect background labeling. However, it inadvertently creates a data imbalance, leading to a bias towards foreground classes. To counteract this, our second loss function is applied to regular images, enhancing the weight on background to neutralize the bias. This dual loss system, which we term balanced anti-misguidance losses, establishes our third balance.

Expanding upon the concepts discussed earlier, we introduce a novel CPS approach named \textbf{Bal}anced \textbf{Con}tinual \textbf{Pa}noptic \textbf{S}egmentation (BalConpas). Our contributions are summarized as follows:
\begin{itemize}
\item We present BalConpas, a new CPS framework distinguished by three key balances. Our experimental results confirm that BalConpas achieves state-of-the-art performance not only in CPS but also in continual semantic and instance segmentation.
\item We propose a past-class backtrace distillation, which selectively distills features associated with previous classes, striking a balance between stability and adaptability.
\item We devise a class-proportional memory strategy and balanced anti-misguidance losses for sample replay. The former creates a replay sample set that reflects a true representation of class balance, enhancing the recall of past knowledge. The latter resolves the adverse effects of incomplete annotations while avoiding classification bias.
\end{itemize}

\section{Related Work}
\label{sec:related_works}

\subsection{Continual Learning}
Continual learning research \cite{li2017learning, ostapenko2019learning, douillard2022dytox, huang2024learning} in the field of deep learning \cite{deng2009imagenet, he2016deep, chen2023kepsalinst, cong2024query} primarily aims to enable neural networks to sequentially acquire knowledge without forgetting previously learned information. This research is most commonly applied to image classification. The main techniques employed can be categorized into three groups: regularization, replay, and dynamic structure. Regularization-based methods \cite{li2017learning, chaudhry2018riemannian, dhar2019learning, douillard2020podnet} focus on constraining model parameter updates to prevent the forgetting of past knowledge. Replay techniques \cite{rebuffi2017icarl, shin2017continual, ostapenko2019learning} involve saving a selection of training samples from previous steps or sourcing similar samples from outside the dataset, then replaying them to reinforce established knowledge. Dynamic structure strategies \cite{mallya2018packnet, mallya2018piggyback, yan2021dynamically, douillard2022dytox} allow models to incorporate new knowledge by adding new parameters while retaining existing ones, ensuring competency across all learned information.

\subsection{Continual Image Segmentation}

The majority of research in continual image segmentation has centered on CSS \cite{michieli2019incremental, cermelli2020modeling, douillard2021plop, michieli2021continual, cha2021ssul, maracani2021recall, zhang2022representation, zhang2022mining, yang2023uncertainty, xiao2023endpoints, chen2023saving}. MiB \cite{cermelli2020modeling} is a forerunner in addressing background shift by manipulating output probabilities. PLOP \cite{douillard2021plop} employs pseudo-labels generated by the prior model and multi-scale local distillation to preserve past knowledge. SSUL \cite{cha2021ssul} introduces an \textit{unknown} class to effectively manage background shifts and creates a replay sample set with an equal number of images per class. RCIL \cite{zhang2022representation} adopts dual branches for accommodating old and new information, coupled with two average-pooling-based distillations for enhanced knowledge retention. EWF \cite{xiao2023endpoints} dynamically fuses models containing past and new knowledge, thereby strengthening memory retention for both. Beyond CSS, there have been strides in continual instance segmentation (CIS) and CPS. In CIS, MTN \cite{gu2021class} employs two teacher networks for past and new knowledge to instruct the student network. In CPS, CoMFormer \cite{cermelli2023comformer} stands as a pioneering work, utilizing adaptive distillation loss and mask-based pseudo-labeling to counteract catastrophic forgetting and background shift.

In our research, we present BalConpas, a framework designed for CPS, but also applicable to CSS and CIS. By integrating three crucial balances, BalConpas demonstrates improvements over existing methods.

\section{Proposed Method}

\subsection{Problem Definition}

In continual segmentation, a model undergoes incremental training across $T$ steps. At each step $t\in \{ 1, \dots ,T \}$, the training set $D^t$ comprises a series of image-label pairs. The label of each image includes $N^{gt}$ ground-truth segments, denoted by $z^{gt}=\{(c_i^{gt}, m_i^{gt})|c_i^{gt}\in \mathcal{C}^t, m_i^{gt}\in \{0,1\}^{H\times W}\}_{i=1}^{N^{gt}}$. Here, $c_i^{gt}$ is the ground-truth class, $m_i^{gt}$ is the ground-truth binary mask, with $H\times W$ indicating the spatial dimensions of the image. $\mathcal{C}^t$ represents the set of classes focused at step $t$. Note that, the sets of classes for different steps are disjoint. The goal after training at step $t$ is to enable the model to accurately predict segments for all seen classes, denoted as $\mathcal{C}^{1:t}$.

\subsection{Overview of the Proposed Method}

\begin{figure*}[!t]
\centering
\includegraphics[width=\textwidth]{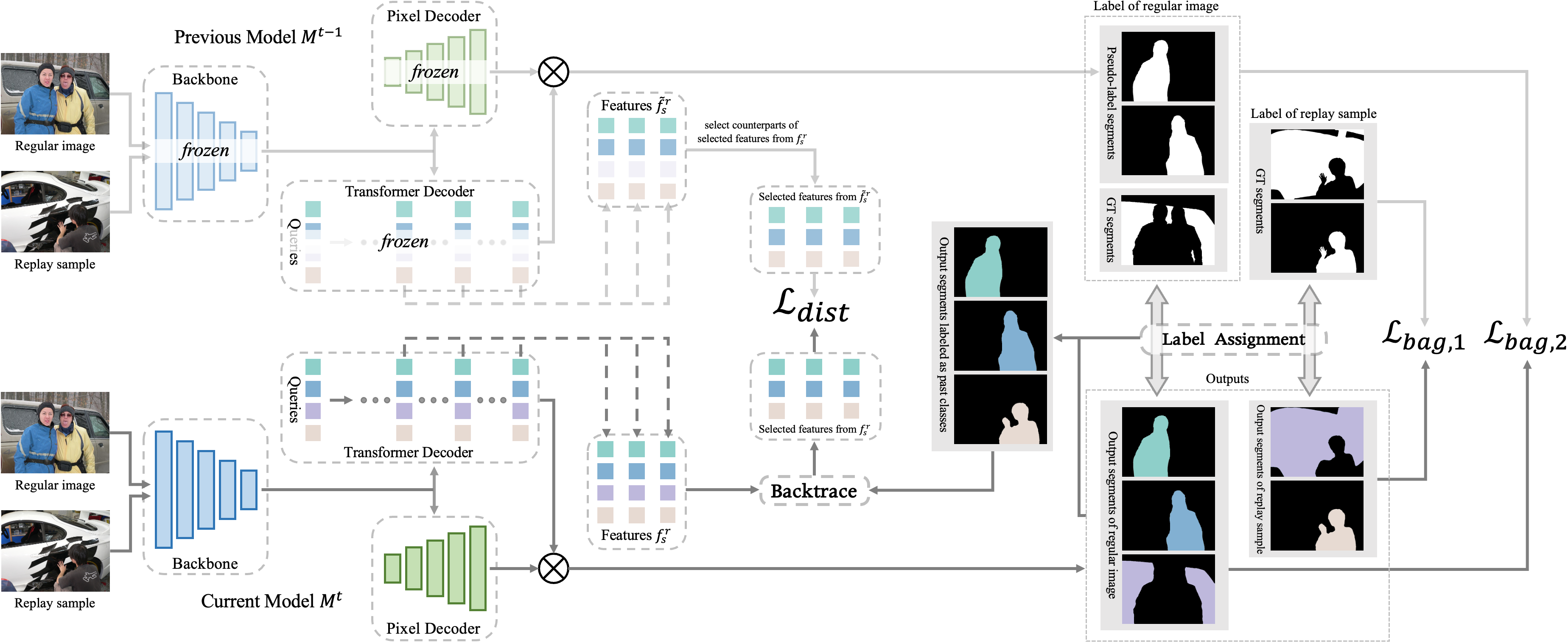}
\caption{Overview of the proposed BalConpas. Given input containing regular images and replay samples, the current and the previous models process it simultaneously. For regular images, outputs from the previous model serve as pseudo-labels, supplementing past-class annotations. After label assignment, we trace back to the features associated with output segments labeled as past classes and focus the knowledge distillation on them. Concurrently, the supervision for replay samples and regular images is managed by the first and second components of the balanced anti-misguidance losses, respectively.}
\label{fig:overview}
\end{figure*}

The proposed BalConpas, illustrated in Fig. \ref{fig:overview}, is built upon the widely-used image segmentation framework, Mask2Former \cite{cheng2022masked}. Its input includes both regular images in $D^t$ and replay samples retained by our class-proportional memory strategy. Given a batch, the previous model, $M^{t-1}$, and the current model, $M^t$, operate in parallel, with the parameters of $M^{t-1}$ being frozen. Following \cite{douillard2021plop, cha2021ssul, cermelli2023comformer}, the predictions of $M^{t-1}$ serve as pseudo-labels of past classes for regular images, supplementing the annotations in $D^t$. After label assignment to output segments (achieved by bipartite matching \cite{carion2020end} in Mask2Former), we backtrack to the features in the transformer decoder associated with segments labeled as past classes. Focusing on these features, we conduct knowledge distillation from $M^{t-1}$ to $M^t$. Finally, the supervision of replay samples and regular images is managed by the first and second components of our balanced anti-misguidance losses, respectively.

\subsection{Past-Class Backtrace Distillation}

In all continual learning scenarios, a paramount challenge is balancing the stability of prior knowledge with the adaptability to new information. Existing continual segmentation methods \cite{michieli2019incremental, douillard2021plop, zhang2022representation} utilize knowledge distillation to restrict changes in entire features or outputs. This constraint, while preserving prior knowledge, may compromise the ability to assimilate new information. We posit that an optimal harmony between stability and adaptability can be realized by pinpointing specific features associated with prior knowledge. By constraining only these identified features and allowing others to evolve freely, the network's capacity to integrate both past and new knowledge can be fully exploited. Inspired by this perspective, we present the past-class backtrace distillation strategy.

In the network, features merge content from past and new classes, making it challenging to identify the parts associated with the past classes. However, towards the end of the workflow, during label assignment, each output segment is matched with an appropriate label. In our base model \cite{cheng2022masked}, this assignment is conducted using similarity-based bipartite matching \cite{carion2020end}. Hence, output segments resembling past-class annotations (sourced from pseudo-labels or replay samples) receive past-class labels, while those similar to new-class annotations are labeled accordingly. This process enables the recognition of segments corresponding to past classes. By retracing the lineage of these segments, we can pinpoint the features related to past classes.

Specifically, the output for each input image comprises a set of $N_q$ segments, denoted by $z=\{(c_i, m_i)\}_{i=1}^{N_q}$, where $c_i$ denotes the classification prediction and $m_i$ represents the mask prediction. $N_q$ is the total number of queries, and hence the number of output segments. As per the workflow:
\begin{equation}
    c_i = \underset{c\in \mathcal{C}^{1:t}}{\arg\max}\;{\rm MLP}(f^r_S[i])[c],
\end{equation}
\begin{equation}
    m_i = {\rm sigmoid}({\rm MLP}(f^r_S[i])\times {\rm MLP}(f^p_S)),
\end{equation}
where $f^r_S$ and $f^p_S$ signify the output features from the transformer decoder and the pixel decoder, respectively. ${\rm MLP}$ denotes a multilayer perceptron, ${\rm sigmoid}$ is the sigmoid function, and $[\cdot]$ indicates the channel or element index.
The above equations imply that if the $i$-th output segment is assigned a past-class label, then the $i$-th channel of $f^r_S$ is specifically associated with that past class. 

As the features output by the transformer decoder, $f^r_S$ are generated by sequentially applying attention layers to the initial queries $f^r_0$ and the backbone features $f^b$. Each attention layer operates according to $f^r_s = {\rm Attn}(f^r_{s - 1}, f^b)$, where $s\in \left[1,S\right]$ indexes the layers, and ${\rm Attn}$ represents the attention operation. Given the one-to-one correspondence between the input and output channels of the attention layer, for all $s \in \left[1,S\right]$, the corresponding $f^r_s[i]$ is associated with past classes if the $i$-th output segment is labeled as such.

Building upon the aforementioned backtrace procedure, we can identify all features associated with past classes after obtaining the label assignment results. Subsequently, we apply knowledge distillation from the previous model to the current model on these identified features, yielding our distillation loss $\mathcal{L}_{dist}$:
\begin{equation}
    \mathcal{L}_{dist} = \sideset{}{_{s=1}^{S}}\sum \mathbbm{1}_{ (c_i, m_i) \in \mathcal{E}_{past} } \, {\rm MSE}(f_s^r[i], \tilde{f}_s^r[i]).
\end{equation}
Here, $\mathbbm{1}$ denotes the indicator function, $\mathcal{E}_{past}$ refers to the set of output segments assigned past-class labels, ${\rm MSE}$ represents the mean square error, and $\tilde{f}_s^r$ indicates the transformer features from the previous model $M^{t-1}$. This distillation focuses on features related to past classes, achieving a balance between stability and adaptability.

\subsection{Class-Proportional Memory}

Existing continual learning research has validated that replaying some past-class samples is effective in preventing catastrophic forgetting. In addition, by enhancing the network’s ability to distinguish between past and new classes, these replay samples also help improve performance on new classes. However, due to the limited capacity of the replay sample set, the selection of samples greatly affects the effectiveness. This brings us to the second key balance: the class balance within the replay sample set. We argue that this balance should not be a mere average but should instead mirror the cumulative class distribution from past training sets. This is because classes that are more common in the training set often exhibit greater diversity, making it more challenging to comprehensively understand their representations. Additionally, classes that frequently appear in the dataset are usually more common in relevant application scenarios. Therefore, incorporating more samples from these classes into the replay sample set is both crucial and logical. Furthermore, such alignment of class distributions is instrumental in maintaining consistent classification tendencies regarding past classes. Based on these considerations, we devise the class-proportional memory strategy. 

Specifically, in the first training step, we compute the occurrence count of ground-truth segments from each current class $c \in \mathcal{C}^1$, denoted by $\pi_c$. These counts enable us to calculate the class distribution $\Pi^1$ as $\Pi^1=\{\frac{\pi_c}{\sum_{c\in \mathcal{C}^{1}}{\pi_c}}\}_{c \in \mathcal{C}^{1}}$. Leveraging this distribution, we construct a sample set $\mathcal{R}^1$ comprising $N_r$ samples. In simple words, the goal is to select $N_r$ images from the current training set $D^1$ that best approximate the desired class distribution $\Pi^1$. Due to the computational complexity of identifying a globally optimal solution for this task, we employ a greedy algorithm to obtain a satisfactory local solution efficiently. Specifically, the greedy algorithm iterates $N_r$ times through all images in the randomly ordered $D^1$, each time selecting an image that maximally narrows the discrepancy between the evolving class distribution and $\Pi^1$. Owing to space constraints, we will provide a detailed description of this algorithm in the appendix. This process can be succinctly expressed by the equation:
\begin{equation}
    \mathcal{R}^1 = \Omega(D^1,N_r,\Pi^1),
\end{equation}
where $\Omega$ denotes the greedy algorithm.

In each subsequent step $t$, we compute the occurrence count for $c \in \mathcal{C}^t$ and update the cumulative class distribution to $\Pi^{t}=\{\frac{\pi_c}{\sum_{c\in \mathcal{C}^{1:t}}{\pi_c}}\}_{c \in \mathcal{C}^{1:t}}$. Utilizing this, we update the replay sample set. However, a challenge arises as we select images from $\mathcal{R}^{t-1} \cup D^t$, with $D^t$ typically containing far more images than $\mathcal{R}^{t-1}$. Owing to the local rather than global optimization of the greedy algorithm, there is a tendency to choose fewer images from $\mathcal{R}^{t-1}$. Hence, after several iterations of updating the replay sample set, there could be a scarcity of images from classes encountered early in the sequence. To address this, we introduce a constraint to preserve a proportion $\lambda^t = \frac{|\mathcal{C}^{1:t-1}|}{|\mathcal{C}^{1:t}|}$ of images from $\mathcal{R}^{t-1}$ when forming $\mathcal{R}^t$, thus mitigate the risk of excessively diluting the representation of classes introduced earlier in the sequence. Here, $|\cdot|$ signifies set cardinality. The update mechanism for the replay sample set is formalized as:
\begin{equation}
\mathcal{R}^t = \Omega(\mathcal{R}^{t-1},\lambda^t N_r,\Pi^t)
\cup \Omega(D^t, (1-\lambda^t) N_r, \Pi^t - \widetilde{\Pi}^t),
\end{equation}
where $\widetilde{\Pi}^t$ represent the class distribution of the first term, $\Omega(\mathcal{R}^{t-1},\lambda^t N_r,\Pi^t)$, namely the provisional class distribution after selecting $\lambda^t N_r$ images from $\mathcal{R}^{t-1}$.

During each incremental training step $t$, we combine $\mathcal{R}^{t-1}$ with $D^t$ for training. The class-proportional selection of replay samples plays a crucial role in preserving past-class knowledge and aiding in distinguishing between new and past classes. Within the size limit of $N_r$, our approach prioritizes classes that are significant in practical applications and more likely to show broad diversity. This maximizes the recall value of the replay sample set and ensures stable classification tendencies. Furthermore, by calculating class distributions based on segment counts rather than image counts per class, our strategy aligns better with the instance-aware requirements of CPS.

\subsection{Balanced Anti-Misguidance Losses}

Despite obtaining a suitable replay sample set, the replay process poses a unique challenge in continual segmentation tasks. Replay samples are annotated solely for the classes from their original step, but they may also contain other past or current classes that lack annotations. During loss computation, these unannotated classes can only be treated as background, also referred to as ``no object'', which can mislead the learning process. To tackle this, we devise balanced anti-misguidance losses. This system consists of two components, each specifically managing the supervision of replay samples and regular images, respectively.

The first component operates on replay samples. It is a modified version of cross-entropy that computes the loss solely for foreground class labels, skipping those labeled as ``no object''. This is encapsulated by the following equation:
\begin{equation}
    \mathcal{L}_{bag,1} = \frac{1}{N_q}\sideset{}{_{i=1}^{N_q}}\sum \mathbbm{1}_{\overline{c}_i^{gt} \neq \varnothing}  \sideset{}{_{j=1}^{\mathcal{C}^{1:t}}}\sum \overline{c}_{i,j}^{gt} \log p_{i,j}.
\end{equation}
Here, $\overline{c}_i^{gt} \in \mathbb{R}^{|\mathcal{C}^{1:t}|}$ denotes the one-hot encoding of $c_i^{gt}$ after it has been reordered based on label assignment results and padded with ``no object'' ($\varnothing$) in empty positions. $j$ is the index, with $\overline{c}_{i,j}^{gt}$ can be $0$ or $1$. $p_{i,j}$ denotes the probability of predicting the $i$-th output segment as the $j$-th class. This loss function avoids erroneously guiding the model to classify certain foreground classes as ``no object'' due to the absence of their annotations in the replay samples.

However, the exclusive focus on foreground classes in the first component implicitly reduces the sample count for the ``no object'' class, creating a data imbalance that may cause a classification bias favoring foreground classes. To solve this, we devise a second component applied to regular images:
\begin{equation}
\begin{split}
\mathcal{L}_{bag,2} = & \frac{1}{N_q}\sideset{}{_{i=1}^{N_q}}\sum (\mathbbm{1}_{\overline{c}_i^{gt} \neq \varnothing}  \sideset{}{_{j=1}^{\mathcal{C}^{1:t}}}\sum \overline{c}_{i,j}^{gt} \log p_{i,j} \\ 
& + \frac{N_{no}^r+N_{no}^g}{N_{no}^g} \mathbbm{1}_{\overline{c}_i^{gt} = \varnothing}  \sideset{}{_{j=1}^{\mathcal{C}^{1:t}}}\sum \overline{c}_{i,j}^{gt} \log p_{i,j}).
\end{split}
\end{equation}
Here, $N_{no}^r$ refers to the number of output segments labeled as ``no object'' from the replay samples within the same batch, while $N_{no}^g$ denotes the number of those from the regular images within this batch. This second component increases the weight of the ``no object'' class in the outputs of regular images to compensate for the underrepresentation in the first component's handling of replay samples, thus neutralizing the classification bias. Essentially, by synergizing these two components, we sidestep problems arising from incomplete annotations in replay samples without incurring classification bias, achieving our third balance.

\subsection{Overall Training Loss}
In conclusion, the total training loss during incremental steps is defined as:
\begin{equation}
    \mathcal{L}_{total} = \alpha(\mathcal{L}_{bag,1} + \mathcal{L}_{bag,2}) + \beta\mathcal{L}_{mask} + \gamma\mathcal{L}_{dist},
\end{equation}
where $\mathcal{L}_{mask}$ represents the mask loss, as defined in \cite{cheng2022masked}. $\alpha$, $\beta$, and $\gamma$ serve as balancing hyper-parameters. Following the weighting of classification and mask losses in \cite{cheng2022masked}, $\alpha$ and $\beta$ are set to $2$ and $5$, respectively. For $\gamma$, we set it empirically to $5$. In the first step, $\mathcal{L}_{bag,1} + \mathcal{L}_{bag,2}$ reduces to a standard cross-entropy loss.

\section{Experiments}
\label{sec:experiments}

\subsection{Experimental Setup}

\subsubsection{Datasets and Evaluation Metric.}
Following \cite{cermelli2023comformer}, we assess our approach using the ADE20K dataset \cite{zhou2017scene}. It includes 20,210 training images and 2,000 validation images, distributed across 150 classes, of which 100 are ``thing'' classes and 50 are ``stuff'' classes. For the three continual segmentation tasks, we employ the respective standard evaluation metrics. In particular, we use Panoptic Quality (PQ) \cite{kirillov2019panoptic} for CPS, mean Intersection over Union (mIoU) for CSS, and Average Precision (AP) \cite{lin2014microsoft} for CIS. Results are reported for base classes $\mathcal{C}^1$ (\textit{base}), incremental classes $\mathcal{C}^{2:T}$ (\textit{inc.}), and all classes (\textit{all}), alongside an average of the outcomes for all seen classes after each step (\textit{avg}).

\subsubsection{Continual Learning Protocols.}
Following existing continual segmentation works \cite{cermelli2020modeling, douillard2021plop, cermelli2023comformer}, we evaluate our model across different class splits over multiple steps. Each split is expressed as $N_1\text{-}N_2$, where $N_1$ denotes the number of classes in the initial step, and $N_2$ indicates the classes in each incremental step. For CPS and CSS, we employ splits of \textit{100-50}, \textit{100-10}, \textit{100-5}, and \textit{50-50}. For CIS, considering only the 100 ``thing'' classes possess instance-level annotations, we restrict our evaluations to the \textit{50-50}, \textit{50-10}, and \textit{50-5} splits. Additionally, we apply the widely-adopted \textit{overlapped} setting in continual segmentation, where an image might appear in multiple steps, but with different annotations corresponding to the focused classes at each respective step.
Due to limited space, the results for CIS and the \textit{50-50} class split for CPS and CSS are provided in the appendix.

\subsubsection{Implementation Details.}
We adopt Mask2Former \cite{cheng2022masked} as our base model. Following previous works \cite{douillard2021plop, gu2021class, cermelli2023comformer}, we utilize an ImageNet \cite{deng2009imagenet} pre-trained ResNet-50 backbone \cite{he2016deep} for both CPS and CIS and an ImageNet \cite{deng2009imagenet} pre-trained ResNet-101 backbone \cite{he2016deep} for CSS. Additionally, the input image resolution is $640\times640$ for CPS and CIS, and $512\times512$ for CSS. We optimize the model using the AdamW optimizer \cite{loshchilov2018decoupled} with an initial learning rate of 0.0001 for the first training step and 0.00005 for incremental steps, with all steps using a batch size of 8. Moreover, training is conducted for 160,000 iterations in the first step, and 1,000 iterations for each class in incremental steps (\eg, training for 10 classes would involve 10,000 iterations). All other hyperparameters are kept at their default values as specified in Mask2Former. In experiments involving replay, we consistently use 300 replay samples, aligning with \cite{cha2021ssul, baek2022decomposed}.

\subsection{Comparisons}

In this part, we benchmark our BalConpas framework against state-of-the-art methods on CPS \cite{cermelli2023comformer} and CSS \cite{cermelli2020modeling, douillard2021plop, xiao2023endpoints, cermelli2023comformer}. Compared methods include MiB \cite{cermelli2020modeling}, PLOP \cite{douillard2021plop}, SSUL \cite{cha2021ssul}, EWF \cite{xiao2023endpoints}, and CoMFormer \cite{cermelli2023comformer}. For a fair comparison, we report the performance metrics for both their original implementations and adaptations to Mask2Former. 

\begin{table*}[t]
  \caption{CPS results on the ADE20K dataset, measured by PQ.}
  \renewcommand{\arraystretch}{1}
  \scriptsize
  \centering
  \begin{tabular}{p{2cm}<{\centering}|p{0.7cm}<{\centering}p{0.7cm}<{\centering}p{0.7cm}<{\centering}p{0.7cm}<{\centering}|p{0.7cm}<{\centering}p{0.7cm}<{\centering}p{0.7cm}<{\centering}p{0.7cm}<{\centering}|p{0.7cm}<{\centering}p{0.7cm}<{\centering}p{0.7cm}<{\centering}p{0.7cm}<{\centering}}
    \toprule
    \multirow{2}*{Model} & \multicolumn{4}{c}{\textbf{100-50 (2 steps)}} & \multicolumn{4}{c}{\textbf{100-10 (6 steps)}} & \multicolumn{4}{c}{\textbf{100-5 (11 steps)}} \\
    & \textit{base} & \textit{inc.} & \textit{all} & \textit{avg} & \textit{base} & \textit{inc.} & \textit{all} & \textit{avg} & \textit{base} & \textit{inc.} & \textit{all} & \textit{avg} \\
    \midrule
    FT & 0.0 & 32.4 & 10.8 & 26.8 & 0.0 & 4.8 & 1.6 & 8.9 & 0.0 & 2.2 & 0.7 & 4.7 \\
    MiB \cite{cermelli2020modeling} & 23.3 & 14.9 & 20.5 & 31.7 & 6.8 & 0.2 & 4.6 & 19.1 & 2.3 & 0.0 & 1.5 & 13.4 \\
    PLOP \cite{douillard2021plop} & 42.4 & 23.7 & 36.2 & 39.5 & 37.7 & 23.3 & 32.9 & 37.8 & 31.1 & 11.9 & 24.7 & 31.3 \\
    SSUL \cite{cha2021ssul} & 35.9 & 18.1 & 30.0 & 33.8 & 31.6 & 11.9 & 25.0 & 30.3 & 30.2 & 7.9 & 22.8 & 27.9 \\
    EWF \cite{xiao2023endpoints} & 37.9 & 4.4 & 26.8 & 34.8 & 35.6 & 0.0 & 23.7 & 32.5 & 32.4 & 0.0 & 21.6 & 31.2 \\
    CoMFormer \cite{cermelli2023comformer} & 41.1 & 27.7 & 36.7 & 38.8 & 36.0 & 17.1 & 29.7 & 35.3 & 34.4 & 15.9 & 28.2 & 34.0 \\
    Ours & 42.8 & 25.7 & \underline{37.1} & \underline{40.0} & 40.7 & 22.8 & \underline{34.7} & \underline{38.8} & 36.1 & 20.3 & \underline{30.8} & \underline{35.8} \\
    Ours-R & 43.2 & 26.1 & \textbf{37.5} & \textbf{40.2} & 41.2 & 27.2 & \textbf{36.5} & \textbf{39.7} & 37.4 & 23.1 & \textbf{32.6} & \textbf{37.4} \\
    \hdashline
    Joint & 43.8 & 30.9 & 39.5 & - & 43.8 & 30.9 & 39.5 & - & 43.8 & 30.9 & 39.5 & - \\
    \bottomrule
  \end{tabular}
  \label{tab:quan_cps}
\end{table*}

\begin{table*}[t]
  \caption{CSS results on the ADE20K dataset, measured by mIoU.}
  \renewcommand{\arraystretch}{1}
  \scriptsize
  \centering
  \begin{tabular}{p{2cm}<{\centering}|p{0.7cm}<{\centering}p{0.7cm}<{\centering}p{0.7cm}<{\centering}p{0.7cm}<{\centering}|p{0.7cm}<{\centering}p{0.7cm}<{\centering}p{0.7cm}<{\centering}p{0.7cm}<{\centering}|p{0.7cm}<{\centering}p{0.7cm}<{\centering}p{0.7cm}<{\centering}p{0.7cm}<{\centering}}
    \toprule
    \multirow{2}*{Model} & \multicolumn{4}{c}{\textbf{100-50 (2 steps)}} & \multicolumn{4}{c}{\textbf{100-10 (6 steps)}} & \multicolumn{4}{c}{\textbf{100-5 (11 steps)}} \\
    & \textit{base} & \textit{inc.} & \textit{all} & \textit{avg} & \textit{base} & \textit{inc.} & \textit{all} & \textit{avg} & \textit{base} & \textit{inc.} & \textit{all} & \textit{avg} \\
    \midrule
    FT & 0.0 & 3.2 & 1.1 & 26.3 & 0.0 & 0.1 & 0.0 & 9.1 & 0.0 & 0.3 & 0.1 & 5.6 \\
    MiB\textsuperscript{$\dagger$} \cite{cermelli2020modeling} & 41.9 & 14.9 & 32.9 & - & 38.2 & 11.1 & 29.2 & - & 36.0 & 5.7 & 26.0 & - \\
    MiB \cite{cermelli2020modeling} & 41.8 & 25.9 & 36.5 & 44.0 & 21.1 & 10.0 & 17.4 & 34.4 & 12.4 & 5.0 & 9.9 & 28.3 \\
    PLOP\textsuperscript{$\dagger$} \cite{douillard2021plop} & 41.9 & 14.9 & 32.9 & 37.4 & 40.5 & 13.6 & 31.6 & 36.6 & 39.1 & 7.8 & 28.8 & 35.3 \\
    PLOP \cite{douillard2021plop} & 49.1 & 29.8 & 42.6 & 47.1 & 44.7 & 23.1 & 37.5 & 43.0 & 35.5 & 17.3 & 29.4 & 36.5 \\
    SSUL\textsuperscript{$\dagger$} \cite{cha2021ssul} & 42.8 & 17.5 & 34.4 & - & 42.9 & 17.7 & 34.5 & - & 42.9 & 17.8 & 34.6 & - \\
    SSUL \cite{cha2021ssul} & 41.7 & 21.6 & 35.0 & 39.6 & 38.5 & 13.7 & 30.2 & 36.2 & 36.9 & 12.7 & 28.8 & 34.8 \\
    EWF\textsuperscript{$\dagger$} \cite{xiao2023endpoints} & 41.2 & 21.3 & 34.6 & - & 41.5 & 16.3 & 33.2 & - & 41.4 & 13.4 & 32.1 & - \\
    EWF \cite{xiao2023endpoints} & 49.2 & 23.7 & 40.7 & 46.1 & 48.5 & 17.7 & 38.2 & 44.5 & 46.9 & 14.7 & \underline{36.2} & \textbf{43.4} \\
    CoMFormer \cite{cermelli2023comformer} & 44.7 & 26.2 & 38.4 & 41.2 & 40.6 & 15.6 & 32.3 & 37.4 & 39.5 & 13.6 & 30.9 & 36.5 \\
    Ours & 49.9 & 30.1 & \underline{43.3} & \underline{47.4} & 47.3 & 24.2 & \underline{38.6} & \underline{43.6} & 42.1 & 17.2 & 33.8 & 41.3 \\
    Ours-R & 50.8 & 30.4 & \textbf{44.0} & \textbf{47.7} & 48.1 & 25.3 & \textbf{40.5} & \textbf{45.4} & 43.9 & 22.7 & \textbf{36.9} & \underline{43.1} \\
    \hdashline
    Joint & 51.7 & 40.2 & 47.8 & - & 51.7 & 40.2 & 47.8 & - & 51.7 & 40.2 & 47.8 & - \\
    \bottomrule
  \end{tabular}
  \label{tab:quan_css}
\end{table*}

\begin{figure*}
    \centering
    \includegraphics[width=\textwidth]{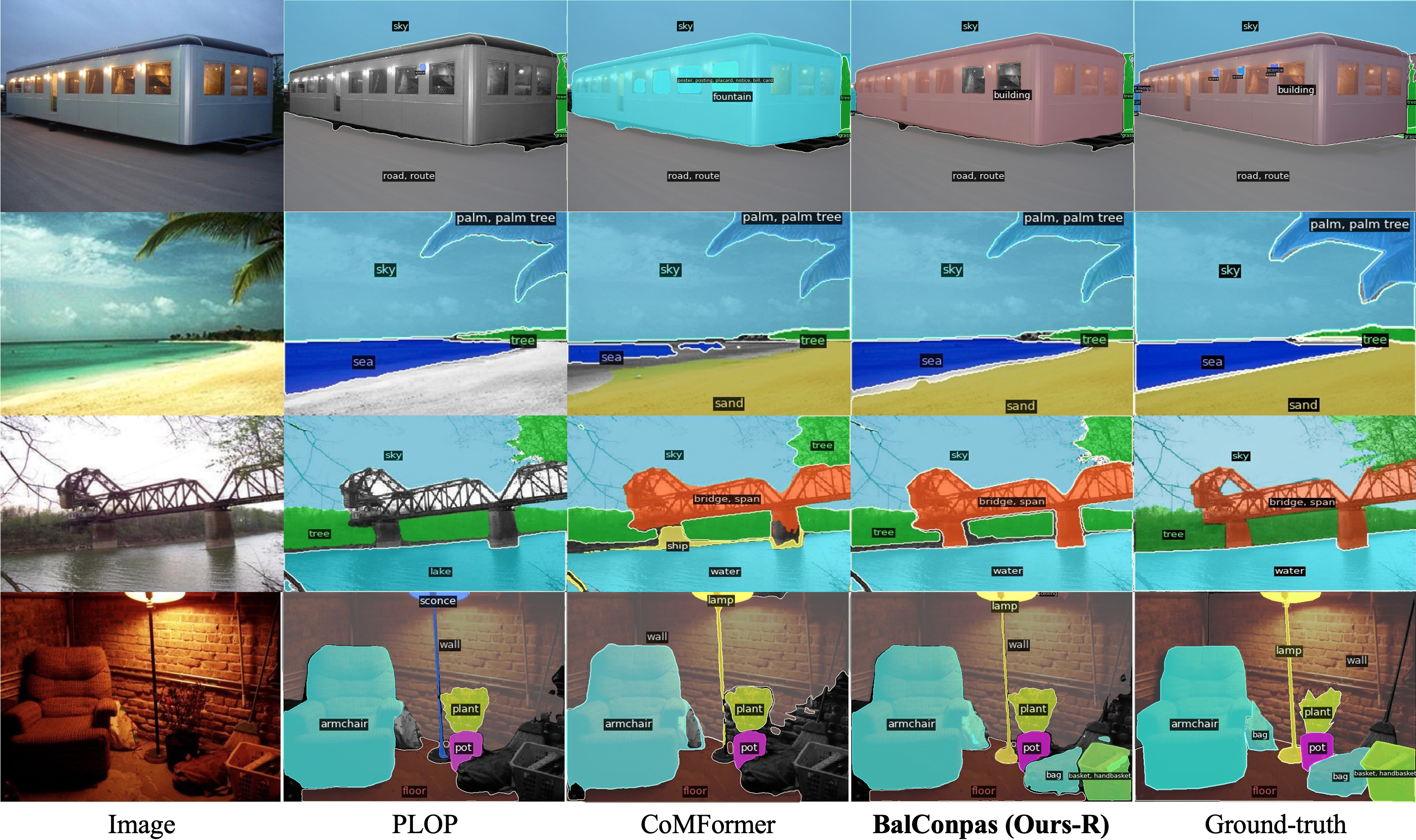}
    \caption{Qualitative comparison of BalConpas and existing methods on ADE20K for \textit{100-10} CPS.}
    \label{fig:qual_comp}
\end{figure*}

\subsubsection{Quantitative comparison.}
The quantitative results for CPS and CSS are shown in Tab. \ref{tab:quan_cps} and Tab. \ref{tab:quan_css}, respectively. We present the performance of our complete model incorporating replay samples (``Ours-R'') and the version without replay samples (``Ours''). For compared models, those labeled with ``$\dagger$'' indicate the original base model implementations, while the others are based on Mask2Former. ``FT'' denotes fine-tuning the base model without employing any continual learning strategies, and ``Joint'' refers to joint training across all data, representing the lower and upper bounds of performance, respectively. 

Tab. \ref{tab:quan_cps} shows that our BalConpas framework surpasses existing methods in all three CPS protocols, especially in the complex \textit{100-10} and \textit{100-5} scenarios. This highlights the superiority of our framework. Notably, even without using replay samples and relying solely on past-class backtrace distillation, our method outperforms all other models, underscoring the effectiveness of this strategy. Methods like MiB \cite{cermelli2020modeling} and EWF \cite{xiao2023endpoints}, targeted at CSS, lack mechanisms for instance-level knowledge and therefore fail completely in CPS. Approaches such as PLOP \cite{douillard2021plop}, SSUL \cite{cha2021ssul}, and CoMFormer \cite{cermelli2023comformer}, despite utilizing pseudo-labels and thereby maintaining some instance-level information, still fall short of the efficacy displayed by our BalConpas. This gap accentuates the significance of our proposed strategies. Our past-class backtrace distillation focuses on features related to individual past-class segments, maintaining instance-level fidelity. Our class-proportional memory strategy, which computes the class distribution among segments, adeptly meets instance-level needs. For ``stuff'' categories that do not differentiate instances, our strategies inherently adjust to focus on the entire category within the image, also yielding favorable results.

Results from Tab. \ref{tab:quan_css} reveal that our BalConpas framework also excels in CSS, surpassing even those methods specifically tailored for this task. Aside from a marginal shortfall in the \textit{avg} metric under \textit{100-5}, where we are slightly outperformed by EWF \cite{xiao2023endpoints}, our framework leads in all other \textit{all} and \textit{avg} metrics. Remarkably, this superior performance is maintained even without incorporating replay samples, solidifying the advantage of our approach over existing methods.

\subsubsection{Qualitative Comparison.}
Fig. \ref{fig:qual_comp} qualitatively compares BalConpas with other methods in the \textit{100-10} CPS scenario of the ADE20K dataset. In the first three examples, PLOP \cite{douillard2021plop} and CoMFormer \cite{cermelli2023comformer} demonstrate varying degrees of forgetting in base classes like \textit{building}, \textit{sand}, \textit{sea}, and \textit{bridge, span}, showing their limited effectiveness in retaining memories of old knowledge. These models also struggle to recognize incremental classes like \textit{bag} and \textit{basket, handbasket} in the last example, with CoMFormer additionally generating false positives for incremental classes \textit{fountain} and \textit{ship} in the first and third examples. These observations underscore the challenges these methods face in learning new knowledge and distinguishing between past and new classes. Conversely, BalConpas effectively addresses these issues. Through past-class backtrace distillation, it balances the retention of old knowledge with the incorporation of new information. It also employs a class-proportional memory strategy and balanced anti-misguidance losses, optimizing the selection and use of replay samples. This not only reinforces the memory of past classes but also aids in differentiating between past and new classes, resulting in BalConpas performing satisfactorily in these examples.

\subsection{Ablation Study}

In this section, we report the results of ablation studies to verify the efficacy of each component and configuration in our BalConpas. For these experiments, we select the \textit{100-10} scenario in CPS, which involves a moderate number of steps, to show performance under average conditions.

\begin{table}[t]
  \caption{Assessment of main components.}
  \renewcommand{\arraystretch}{1}
  \scriptsize
  \centering
  \begin{tabular}{p{1.05cm}<{\centering}|p{1.2cm}<{\centering}p{1.2cm}<{\centering}p{1.2cm}<{\centering}|p{1.05cm}<{\centering}|p{0.75cm}<{\centering}p{0.75cm}<{\centering}p{0.75cm}<{\centering}p{0.75cm}<{\centering}}
    \toprule
    \multirow{2}*{\textit{PCBD}} & \multicolumn{3}{c}{\textit{Replay sample}} & \multirow{2}*{\textit{BAG}} & \multicolumn{4}{c}{\textbf{Panoptic 100-10}} \\
    \cmidrule(lr){2-4}
    & \textit{Random} & \textit{SSUL} & \textit{CPM} &  & \textit{base} & \textit{inc.} & \textit{all} & \textit{avg} \\
    \midrule
     &  &  &  &  & 37.6 & 20.7 & 32.0 & 37.3 \\
    \checkmark &  &  &  &  & 40.7 & 22.8 & 34.7 & 38.8 \\
    \checkmark & \checkmark &  &  & \checkmark & 40.6 & 24.4 & 35.2 & 39.4 \\
    \checkmark &  & \checkmark &  & \checkmark & 40.7 & 26.2 & 35.8 & 39.5 \\
    \checkmark &  &  & \checkmark & \checkmark & 41.2 & 27.2 & \textbf{36.5} & \textbf{39.7} \\
    \checkmark &  &  & \checkmark &  & 40.3 & 24.8 & 35.1 & 39.2 \\
    \bottomrule
  \end{tabular}
  \label{tab:abla_main}
\end{table}

\subsubsection{Main Components.}
Firstly, we assess the impact of each main component: past-class backtrace distillation (\textit{PCBD}), class-proportional memory (\textit{CPM}), and balanced anti-misguidance losses (\textit{BAG}), as detailed in Tab. \ref{tab:abla_main}. The baseline results, shown in the first row, are obtained by applying the pseudo-label strategy \cite{douillard2021plop, cha2021ssul, cermelli2023comformer} to the base model.

The second row of Tab. \ref{tab:abla_main} illustrates that incorporating \textit{PCBD} leads to improvements in the \textit{all} and \textit{avg} metrics of PQ by 2.7 and 1.5, respectively. This strategy targets distillation on features related to past classes, thus preserving performance for these classes without compromising the adaptability to new ones. Note that, the accurate recognition of past and new classes is interdependent: by securing the accuracy of past classes, the model inherently enhances its ability to distinguish between past and new classes, thereby indirectly boosting performance on the latter.

From row three to five, we introduce sample replay in conjunction with \textit{BAG}. Here, \textit{Random} denotes the random selection of replay samples, and \textit{SSUL} represents the sample selection strategy employed in \cite{cha2021ssul,baek2022decomposed}, which selects an equal number of images containing each class. It is evident that all sample selection strategies improve performance compared to not using replay. However, the performance gain brought by our proposed \textit{CPM} notably exceeds that of both \textit{Random} and \textit{SSUL}, affirming the superiority of our class-proportional sample selection strategy. This strategy not only provides the network with more opportunities to revisit classes that are prevalent in previous steps but also helps maintain classification tendencies.

In the sixth row, it can be observed that removing \textit{BAG} noticeably reduces performance. This is because the replay samples contain class annotations only from their original step, potentially causing forgetting of other past classes and impacting the learning of new ones. \textit{BAG} effectively avoids the pitfalls of incomplete annotations and prevents classification bias, thus enhancing performance.

\begin{table}[t]
  \centering
  \caption{Left: Comparison of different distillation strategies. Right: Comparison of different class distribution basis.}
  \begin{minipage}{0.48\linewidth}
    \renewcommand{\arraystretch}{1}
    \scriptsize
    \centering
    \begin{tabular}{p{1.65cm}<{\centering}|p{0.75cm}<{\centering}p{0.75cm}<{\centering}p{0.75cm}<{\centering}p{0.75cm}<{\centering}}
      \toprule
      \multirow{2}*{Distillation} & \multicolumn{4}{c}{\textbf{Panoptic 100-10}} \\
      & \textit{base} & \textit{inc.} & \textit{all} & \textit{avg} \\
      \midrule
      \textit{None} & 37.6 & 20.7 & 32.0 & 37.3 \\
      \textit{Entire} & 39.6 & 22.2 & 33.8 & 38.2 \\
      \textit{PCBD} & 40.7 & 22.8 & \textbf{34.7} & \textbf{38.8} \\
      \bottomrule
    \end{tabular}
    \label{tab:abla_dist}
  \end{minipage}\hfill
  \begin{minipage}{0.48\linewidth}
    \renewcommand{\arraystretch}{1}
    \scriptsize
    \centering
    \begin{tabular}{p{2.3cm}<{\centering}|p{0.75cm}<{\centering}p{0.75cm}<{\centering}p{0.75cm}<{\centering}p{0.75cm}<{\centering}}
      \toprule
      \multirow{2}*{Class Dist. Basis} & \multicolumn{4}{c}{\textbf{Panoptic 100-10}} \\
      & \textit{base} & \textit{inc.} & \textit{all} & \textit{avg} \\
      \midrule
      \textit{Random} & 40.6 & 24.4 & 35.2 & 39.4 \\
      \textit{Pixel} & 40.8 & 24.1 & 35.2 & 39.2 \\
      \textit{Image} & 41.5 & 25.0 & 36.0 & 39.4 \\
      \textit{Segment (CPM)} & 41.2 & 27.2 & \textbf{36.5} & \textbf{39.7} \\
      \bottomrule
    \end{tabular}
    \label{tab:abla_mem}
  \end{minipage}
\end{table}

\subsubsection{Distillation Strategy.}
In the left part of Tab. \ref{tab:abla_dist}, we compare two distillation strategies: distillation of entire features in $\{f_s^r\}_{s=1}^{S}$ (\textit{Entire}) and our \textit{PCBD}, which selectively distills past class features. Compared to the baseline without knowledge distillation (\textit{None}), both strategies show improvements, but \textit{PCBD} performs better. This is because while distilling entire features may preserve old knowledge, it can impair the ability to integrate new knowledge, leading to decreased performance on new classes. Moreover, the compromised ability to learn new classes undermines the discrimination between new and past classes, indirectly affecting performance on the latter. In contrast, \textit{PCBD} specifically focuses on features related to past knowledge, effectively resolving this issue.

\subsubsection{Class Distribution Basis.}
In the right part of Tab. \ref{tab:abla_mem}, we assess the impact of using different entities as the basis for defining class distribution within \textit{CPM}, including \textit{Pixel}, \textit{Image}, and the default \textit{Segment}. A random selection of replay samples (\textit{Random}) serves as our baseline, with its results presented in the first row. Next, we evaluate the effect of class distribution determined by pixel presence across different classes. As indicated in the second row, this approach yields only marginal improvement over \textit{Random}. Given the mask classification paradigm of our base model, a pixel-centric method does not significantly enhance performance. The results in the third row show that defining class distribution based on the count of images containing different classes offers some enhancement over \textit{Random}. However, this image-centric approach overlooks the possibility of multiple segments from the same class appearing in a single image, rendering it less effective than our segment-centric approach (\textit{Segment}). For CPS, recognizing the potential for multiple segments from the same class within one image is crucial.

\section{Conclusion}
\label{sec:conclusion}

This paper presents BalConpas, a novel CPS method that focuses on three critical balances. Firstly, we introduce past-class backtrace distillation. This technique strikes a balance between preserving existing knowledge and adapting to new information by selectively distilling features related to past classes while allowing other features the flexibility to learn new classes. Secondly, we devise a class-proportional memory strategy for the class balance in the replay sample set. This strategy selects replay samples based on the cumulative class distribution from past training sets, prioritizing significant and challenging classes, and simultaneously ensuring stable classification tendencies. Finally, to tackle the issue of incomplete annotations in replay samples, we propose balanced anti-misguidance losses. This solution effectively mitigates the negative impact of partial annotations without introducing data imbalance, establishing the third balance. Our experimental results demonstrate that BalConpas achieves state-of-the-art performance not only in CPS but also in CSS and CIS.

\section*{Acknowledgements}
This work was supported in part by the National Science and Technology Major Project under Grant 2021ZD0112100, 
in part by the Taishan Scholar Project of Shandong Province under Grant tsqn202306079, and in part by Xiaomi Young Talents Program.

\bibliographystyle{splncs04}
\bibliography{main}

\newpage
\appendix
\renewcommand\thefigure{\Alph{section}\arabic{figure}}
\renewcommand\thetable{\Alph{section}\arabic{table}}
\setcounter{figure}{0} 
\setcounter{table}{0} 

\section{Greedy Algorithm in Class-Proportional Memory Strategy}
\label{sec:greedy}

In this section, we provide a detailed explanation of the greedy algorithm $\Omega$, employed in our class-proportional memory strategy, as delineated in Alg. \ref{alg:greedy}. This algorithm describes the process of constructing the replay sample set at step $t$. The sample pool $P$, the number of samples to be selected $N$, and the desired class distribution $\Pi$ correspond to the three parameters of $\Omega$, as specified in the main text. The outcome of this algorithm is a selected sample set $R$.

\SetKwInput{KwRequire}{Require}
\begin{algorithm}
\scriptsize
\DontPrintSemicolon
\KwRequire{Sample pool \( P=\{ p_k \}_{k=1}^{N_P} \), number of samples to be selected \( N \), desired class distribution \( \Pi \)}
\KwResult{Selected sample set \( R \)}

\( R \leftarrow \) empty set\;
\( \Phi \leftarrow \) zero vector of size \( \mathcal{C}^{1:t} \)\;

\For{\( n \leftarrow 1 \) \KwTo \( N \)}{
    \( d_{\text{best}} \leftarrow \infty \)\;
    
    \For{\( k \leftarrow 1 \) \KwTo \( N_P \)}{
        \If{\( p_k \notin R \)}{
            \( \phi_k \leftarrow \) count of segments per class in \( p_k \)\;
            \( \Phi_{\text{tmp}} \leftarrow \Phi + \phi_k \)\;
            \( \Pi_{\text{tmp}} \leftarrow \frac{\Phi_{\text{tmp}}}{\sum\Phi_{\text{tmp}}} \)\;
            \( d_{\text{tmp}} \leftarrow \sum\left|\Pi - \Pi_{\text{tmp}}\right| \)\;
            
            \If{\( d_{\text{tmp}} < d_{\text{best}} \)}{
                \( d_{\text{best}} \leftarrow d_{\text{tmp}} \)\;
                \( p_{\text{best}} \leftarrow p_k \)\;
                \( \phi_{\text{best}} \leftarrow \phi_k \)\;
            }
        }
    }
    \( R \leftarrow R \cup \{p_{\text{best}}\} \)\;
    \( \Phi \leftarrow \Phi + \phi_{\text{best}} \)\;
}
\KwRet \( R \)\;

\caption{Greedy Algorithm $\Omega$}
\label{alg:greedy}
\end{algorithm}

To commence, we initialize $R$ as an empty set and $\Phi\in \mathbbm{R}^{\mathcal{C}^{1:t}}$, a vector representing the counts of segments per class in $R$, as a zero vector. The algorithm then iterates over each sample in the pool $P$ a total of $N$ times. The goal of each iteration is to select a sample that minimizes the discrepancy between the current class distribution in $R$ and the desired distribution $\Pi$. At the start of each iteration, we assign an initial value of infinity to the variable $d_{best}$, representing the smallest discrepancy achievable in that iteration. For each sample $p_k$ in $P$ that is not already in $R$, we calculate $\phi_k$, the counts of segments per class in $p_k$. Then, we envision $\Phi_{tmp}$, representing the counts of segments per class in $R$ if $p_k$ were added, and calculate the corresponding class distribution $\Pi_{tmp}$. Subsequently, the discrepancy $d_{tmp}$, quantifying the difference between $\Pi_{tmp}$ and $\Pi$, is computed. If $d_{tmp}$ is smaller than the current $d_{best}$, it suggests that adding $p_k$ to $R$ would bring us closer to the desired class distribution. Thus, we update $d_{best}$ to $d_{tmp}$, designate $p_k$ as $p_{best}$, and record its segment counts per class as $\phi_{best}$. At the end of each iteration, the final $p_{best}$ represents the optimal sample to minimize the discrepancy between the class distribution in $R$ and $\Pi$. This sample is then added to $R$, and we update the segment count vector $\Phi$ accordingly.

\section{Additional Experiment Results}
\label{sec:add_experiments}

\subsection{Comparison}

In this section, we conduct a quantitative comparison of our BalConpas framework against prior state-of-the-art methods using the \textit{50-50} protocols of both continual panoptic segmentation (CPS) and continual semantic segmentation (CSS). In addition, we assess the performance of BalConpas relative to existing leading approaches in the continual instance segmentation (CIS) task, employing three different protocols: \textit{50-50}, \textit{50-10}, and \textit{50-5}. Moreover, visual results of BalConpas and previous state-of-the-art methods in CSS and CIS tasks are presented for further analysis.

\subsubsection{Quantitative comparison for 50-50 CPS and 50-50 CSS.}

\begin{table}[t]
  \centering
  \caption{Left: \textit{50-50} CPS results on the ADE20K dataset, measured by PQ. Right: \textit{50-50} CSS results on the ADE20K dataset, measured by mIoU.}
  \begin{minipage}{0.48\linewidth}
    \renewcommand{\arraystretch}{1}
    \scriptsize
    \centering
    \begin{tabular}{p{2cm}<{\centering}|p{0.7cm}<{\centering}p{0.7cm}<{\centering}p{0.7cm}<{\centering}p{0.7cm}<{\centering}}
    \toprule
    \multirow{2}*{Model} & \multicolumn{4}{c}{\textbf{50-50 CPS (3 steps)}} \\
    & \textit{base} & \textit{inc.} & \textit{all} & \textit{avg} \\
    \midrule
    FT & 0.0 & 16.3 & 10.9 & 26.7 \\
    MiB \cite{cermelli2020modeling} & 18.2 & 8.2 & 11.5 & 29.5 \\
    PLOP \cite{douillard2021plop} & 50.7 & 24.5 & 33.2 & 40.9 \\
    SSUL \cite{cha2021ssul} & 45.6 & 20.0 & 28.5 & 36.6 \\
    EWF \cite{xiao2023endpoints} & 35.1 & 2.3 & 13.2 & 29.7 \\
    CoMFormer \cite{cermelli2023comformer} & 45.2 & 26.5 & 32.7 & 37.9 \\
    Ours & 51.2 & 26.5 & \underline{34.7} & \underline{42.0} \\
    Ours-R & 51.2 & 28.1 & \textbf{35.8} & \textbf{42.6} \\
    \hdashline
    \textit{Joint} & 50.7 & 33.9 & 39.5 & - \\
    \bottomrule
    \end{tabular}
  \end{minipage}\hfill
  \begin{minipage}{0.48\linewidth}
    \renewcommand{\arraystretch}{1}
    \scriptsize
    \centering
    \begin{tabular}{p{2cm}<{\centering}|p{0.7cm}<{\centering}p{0.7cm}<{\centering}p{0.7cm}<{\centering}p{1cm}<{\centering}}
    \toprule
    \multirow{2}*{Model} & \multicolumn{4}{c}{\textbf{50-50 CSS (3 steps)}} \\
    & \textit{base} & \textit{inc.} & \textit{all} & \textit{avg} \\
    \midrule
    FT & 0.0 & 1.7 & 1.1 & 21.8 \\
    MiB\textsuperscript{$\dagger$} \cite{cermelli2020modeling} & 45.6 & 21.0 & 29.3 & - \\
    MiB \cite{cermelli2020modeling} & 37.0 & 18.3 & 24.5 & 41.0 \\
    PLOP\textsuperscript{$\dagger$} \cite{douillard2021plop} & 48.8 & 21.0 & 30.4 & 39.4 \\
    PLOP \cite{douillard2021plop} & 54.9 & 30.2 & 38.4 & 48.3 \\
    SSUL\textsuperscript{$\dagger$} \cite{cha2021ssul} & 49.1 & 20.1 & 29.8 & - \\
    SSUL \cite{cha2021ssul} & 51.5 & 27.4 & 35.4 & 45.1 \\
    EWF\textsuperscript{$\dagger$} \cite{douillard2021plop} & 47.2 & 19.6 & 28.8 & 38.5 \\
    EWF \cite{xiao2023endpoints} & 50.6 & 27.0 & 34.9 & 46.5 \\
    CoMFormer \cite{cermelli2023comformer} & 49.2 & 26.6 & 34.1 & 36.6 \\
    Ours & 55.8 & 33.3 & \underline{40.8} & \underline{49.2} \\
    Ours-R & 56.2 & 33.9 & \textbf{41.3} & \textbf{49.7} \\
    \hdashline
    \textit{Joint} & 56.6 & 43.5 & 47.8 & - \\
    \bottomrule
    \end{tabular}
  \end{minipage}
\label{tab:50-50_cps_css}
\end{table}

Tab. \ref{tab:50-50_cps_css} displays the quantitative comparison of our method with previous methods on the \textit{50-50} CPS and \textit{50-50} CSS protocols. The results in these tables indicate that our method maintains its advantages as noted in other protocols detailed in the main text. The two versions of our method, with and without sample replay, rank first and second, respectively. This consistent high performance across different protocols underscores the reliability and effectiveness of our ideas.

\subsubsection{Quantitative comparison for CIS.}

\begin{table*}[t]
  \caption{CIS results on the ADE20K dataset, measured by AP.}
  \renewcommand{\arraystretch}{1}
  \scriptsize
  \centering
  \begin{tabular}{p{2cm}<{\centering}|p{0.7cm}<{\centering}p{0.7cm}<{\centering}p{0.7cm}<{\centering}p{0.7cm}<{\centering}|p{0.7cm}<{\centering}p{0.7cm}<{\centering}p{0.7cm}<{\centering}p{0.7cm}<{\centering}|p{0.7cm}<{\centering}p{0.7cm}<{\centering}p{0.7cm}<{\centering}p{0.7cm}<{\centering}}
    \toprule
    \multirow{2}*{Model} & \multicolumn{4}{c}{\textbf{50-50 (2 steps)}} & \multicolumn{4}{c}{\textbf{50-10 (6 steps)}} & \multicolumn{4}{c}{\textbf{50-5 (11 steps)}} \\
    & \textit{base} & \textit{inc.} & \textit{all} & \textit{avg} & \textit{base} & \textit{inc.} & \textit{all} & \textit{avg} & \textit{base} & \textit{inc.} & \textit{all} & \textit{avg} \\
    \midrule
    FT & 0.0 & 20.0 & 10.0 & 21.6 & 0.0 & 3.6 & 1.8 & 7.6 & 0.0 & 1.5 & 0.8 & 4.3 \\
    MiB \cite{cermelli2020modeling} & 16.7 & 15.4 & 16.0 & 24.6 & 5.0 & 7.3 & 6.2 & 16.0 & 1.0 & 2.0 & 1.5 & 11.3 \\
    MTN \cite{gu2021class} & 25.8 & 16.4 & 21.1 & 27.2 & 0.2 & 10.7 & 5.5 & 16.3 & 0.0 & 5.3 & 2.6 & 11.1 \\
    PLOP \cite{douillard2021plop} & 32.2 & 17.2 & \underline{24.7} & \textbf{29.0} & 30.8 & 14.2 & 22.5 & 27.0 & 27.8 & 10.1 & 18.9 & 25.0 \\
    SSUL \cite{cha2021ssul} & 26.7 & 11.6 & 19.1 & 23.5 & 23.7 & 8.4 & 16.0 & 21.0 & 21.3 & 6.4 & 13.9 & 19.3 \\
    EWF \cite{xiao2023endpoints} & 29.7 & 11.0 & 20.4 & 26.8 & 27.6 & 4.5 & 16.1 & 23.5 & 25.9 & 2.9 & 14.4 & 22.4 \\
    CoMFormer \cite{cermelli2023comformer} & 26.5 & 10.4 & 18.4 & 22.9 & 22.7 & 7.2 & 15.0 & 20.2 & 18.5 & 6.6 & 12.5 & 18.4 \\
    Ours & 32.2 & 16.7 & 24.5 & \underline{28.9} & 31.3 & 14.8 & \underline{23.1} & \underline{27.2} & 28.3 & 12.5 & \underline{20.4} & \underline{25.7} \\
    Ours-R & 32.3 & 17.2 & \textbf{24.8} & \textbf{29.0} & 31.2 & 15.7 & \textbf{23.4} & \textbf{27.3} & 28.5 & 13.9 & \textbf{21.2} & \textbf{26.2} \\
    \hdashline
    \textit{Joint} & 32.3 & 20.8 & 26.6 & - & 32.3 & 20.8 & 26.6 & - & 32.3 & 20.8 & 26.6 & - \\
    \bottomrule
  \end{tabular}
  \label{tab:quan_cis}
\end{table*}

Tab. \ref{tab:quan_cis} provides a quantitative comparison of our method against existing methods across three protocols of the CIS task. Consistent with results in CPS and CSS, our method demonstrates superior performance over existing methods. Collectively, these findings establish BalConpas as a state-of-the-art solution in CPS, CSS, and CIS tasks, validating its efficacy as a universal continual segmentation framework.

\subsubsection{Qualitative comparison for CSS and CIS.}

\begin{figure*}
    \centering
    \includegraphics[width=\textwidth]{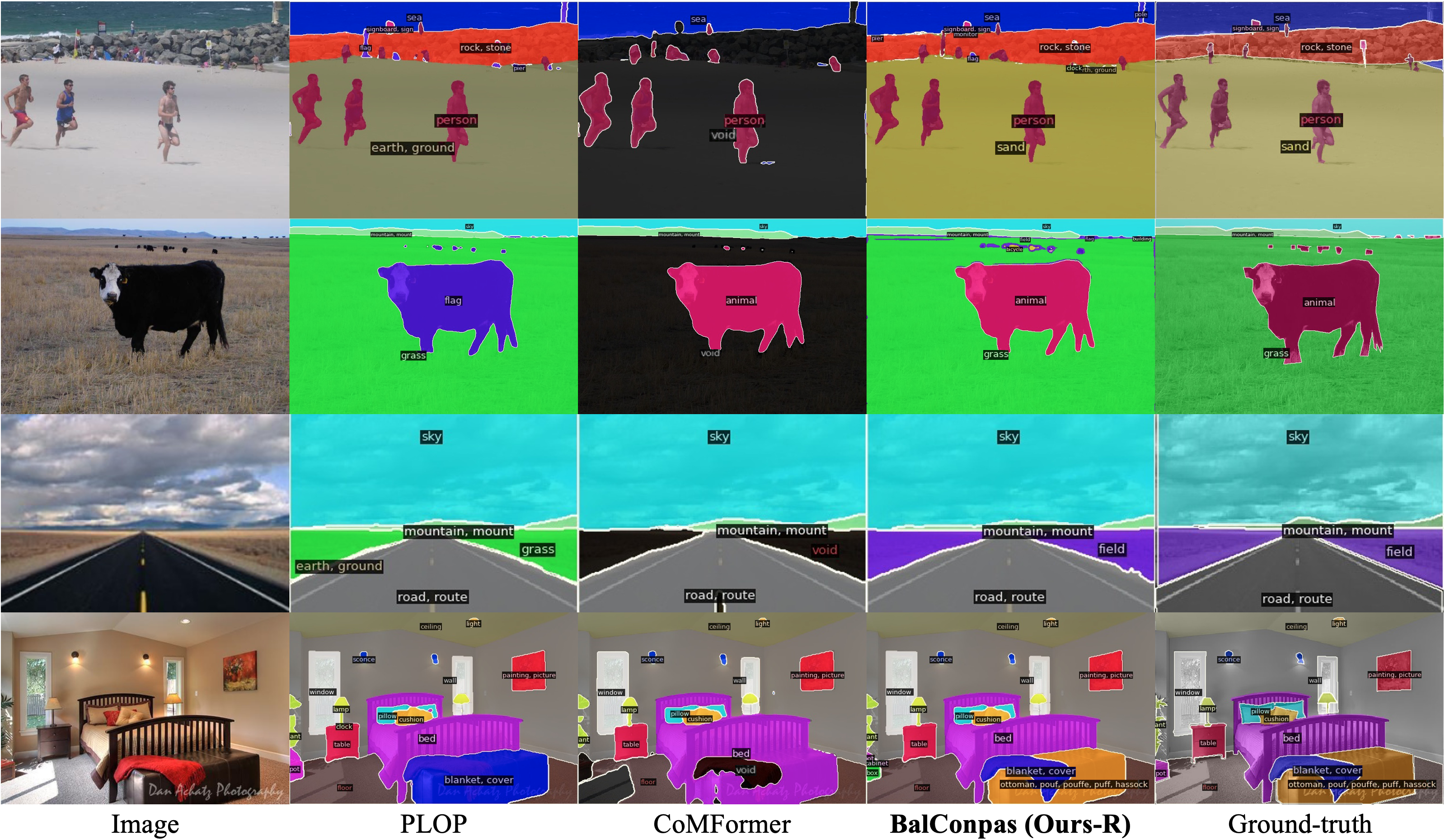}
    \caption{Qualitative comparison of BalConpas and existing methods on ADE20K for \textit{100-10} CSS.}
    \label{fig:qual_comp_css}
\end{figure*}

\begin{figure*}
    \centering
    \includegraphics[width=\textwidth]{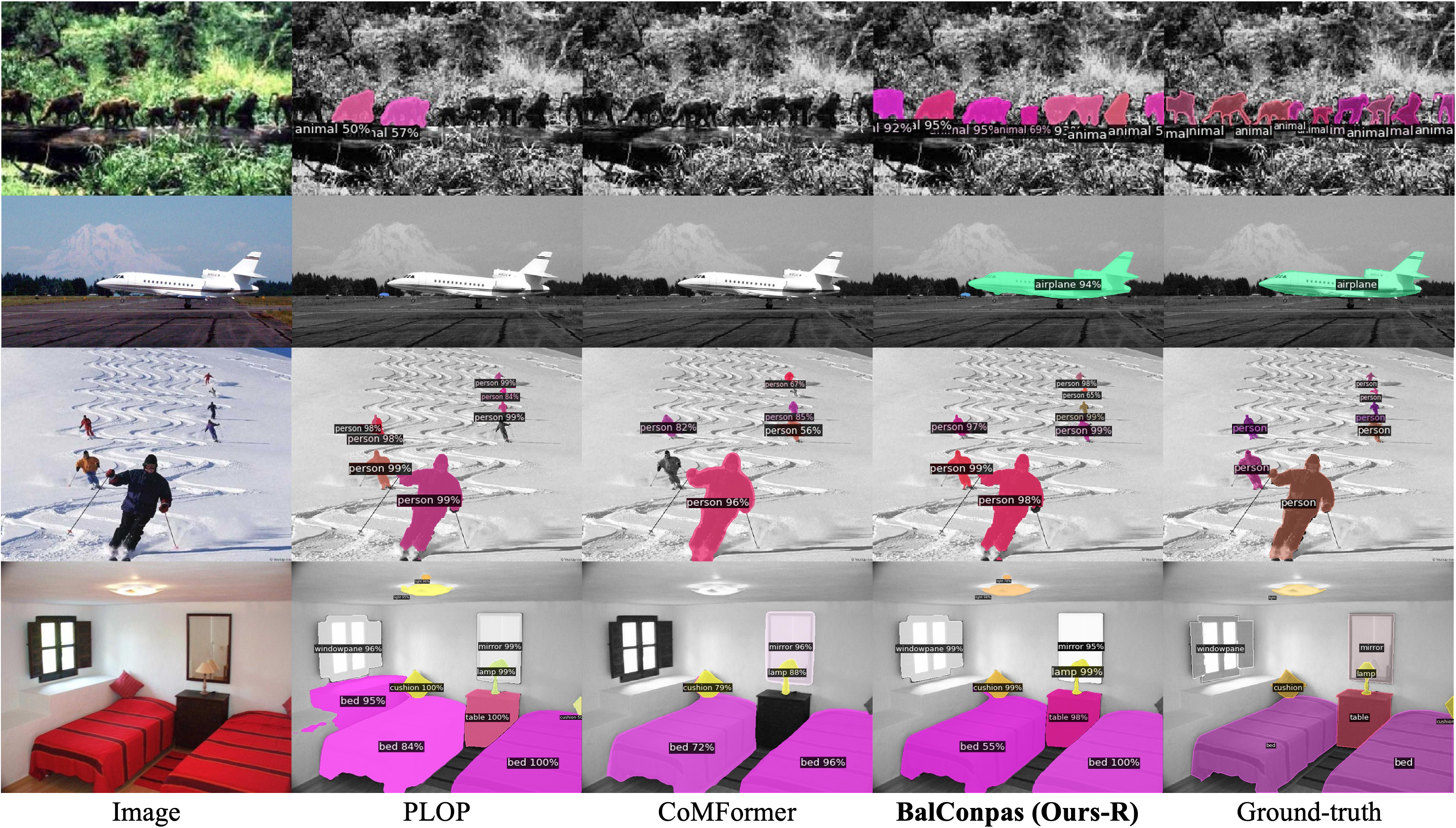}
    \caption{Qualitative comparison of BalConpas and existing methods on ADE20K for \textit{50-10} CIS.}
    \label{fig:qual_comp_cis}
\end{figure*}

Fig. \ref{fig:qual_comp_css} and Fig. \ref{fig:qual_comp_cis} provide qualitative comparisons between BalConpas and state-of-the-art methods for the \textit{100-10} CSS and \textit{50-10} CIS protocols, respectively. In line with our quantitative findings, BalConpas demonstrates satisfactory performance.

In the CSS task, as illustrated in the first example of Fig. \ref{fig:qual_comp_css}, competing methods incorrectly classify the base class \textit{sand} as \textit{earth, ground} or \textit{void} (a class unique to CoMFormer \cite{cermelli2023comformer}, akin to ``no object"). The second example shows these methods either confusing the incremental classes \textit{animal} and \textit{flag} or failing to recognize the base class \textit{grass}. The final two examples highlight challenges with the base classes \textit{field} and \textit{ottoman, pouf, pouffe, puff, hassock}, along with the incremental class \textit{blanket, cover}, where both PLOP \cite{douillard2021plop} and CoMFormer \cite{cermelli2023comformer} struggle. In contrast, our BalConpas accurately predicts both base and incremental classes in these scenarios, emphasizing the effectiveness of the three balances integral to our method. These three balances enable BalConpas to effectively maintain past knowledge while also facilitating improved learning of new knowledge.

For the CIS task, the first example in Fig. \ref{fig:qual_comp_cis} demonstrates the relatively comprehensive identification of multiple instances from the incremental class \textit{animal} by our BalConpas, while the compared methods miss a large portion or all of them. Similarly, the second example highlights the failure of competing methods to detect the incremental class \textit{airplane}, whereas our BalConpas succeeds. The third example shows that PLOP \cite{douillard2021plop} misses the fourth \textit{person} instance from the top on the right and mistakenly splits a single \textit{person} instance on the top left into two, while CoMFormer \cite{cermelli2023comformer} overlooks the second \textit{person} instances from the top on both sides. Conversely, BalConpas accurately identifies all \textit{person} instances. The final example showcases the false positive identification of the base class \textit{bed} by PLOP \cite{douillard2021plop} and the omission of base classes \textit{table} and \textit{windowpane} by CoMFormer \cite{cermelli2023comformer}, whereas our BalConpas adeptly handles them. These observations confirm the superior performance of BalConpas in the CIS task, reiterating the significance of the three balances.

\subsection{Validation of Balanced Anti-Misguidance Losses}

\begin{table}[t]
  \renewcommand{\arraystretch}{1}
  \scriptsize
  \centering
  \caption{Validation of balanced anti-misguidance losses.}
  \begin{tabular}{p{1.5cm}<{\centering}p{1.5cm}<{\centering}|p{0.7cm}<{\centering}p{0.7cm}<{\centering}p{0.7cm}<{\centering}p{0.7cm}<{\centering}}
    \toprule
    \multirow{2}*{\textit{1st comp.}} & \multirow{2}*{\textit{2nd comp.}} & \multicolumn{4}{c}{\textbf{Panoptic 100-10}} \\
     &  & \textit{base} & \textit{inc.} & \textit{all} & \textit{avg} \\
    \midrule
     &  & 40.3 & 24.8 & 35.1 & 39.2 \\
    \checkmark &  & 40.4 & 24.4 & 35.1 & 38.9 \\
    \checkmark & \checkmark & 41.2 & 27.2 & \textbf{36.5} & \textbf{39.7} \\
    \bottomrule
  \end{tabular}
  \label{tab:abla_bag}
\end{table}

In Tab. \ref{tab:abla_bag}, we assess the effectiveness of the two components comprising our balanced anti-misguidance losses. The results indicate that employing only the first component (as shown in the second row) yields no improvement over the setup where neither component is utilized (first row). This outcome arises because the first component exclusively focuses on foreground annotations, effectively mitigating the impact of incorrect ``no object" (background) labels in replay samples but inadvertently causing data imbalance and subsequent classification bias. Notable performance enhancement is observed only upon integrating the second component (third row), which increases the weight of the ``no object" class in regular images. This adjustment compensates for the deficiencies of the first component, achieving data balance. These findings underscore the criticality of maintaining this balance.

\end{document}